\documentclass[conference]{IEEEtran}
\IEEEoverridecommandlockouts

\usepackage{amsmath,amssymb,amsfonts}
\usepackage{algorithmicx}
\usepackage{algpseudocode}
\usepackage{graphicx}
\usepackage{textcomp}
\usepackage{xcolor}
\usepackage{algorithm}
\usepackage{algpseudocode}
\usepackage{tcolorbox}
\usepackage[table]{xcolor}
\usepackage{multirow}
\tcbuselibrary{breakable}
\usepackage{newfloat}
\usepackage{listings}
\usepackage{tikz}
\usepackage{pgfplots}
\usepackage{amsmath}
\usepackage{booktabs}

\def\BibTeX{{\rm B\kern-.05em{\sc i\kern-.025em b}\kern-.08em
    T\kern-.1667em\lower.7ex\hbox{E}\kern-.125emX}}
\begin{document}


\title{CRAFT: Fine-Tuning Pre-hoc Explainability in AI-native 6G RAN}

\author{\IEEEauthorblockN{1\textsuperscript{st} Pranshav Gajjar}
\IEEEauthorblockA{\textit{NextG Wireless Lab} \\
\textit{NCSU}\\
Raleigh USA}
\and
\IEEEauthorblockN{2\textsuperscript{nd} Vijay K Shah}
\IEEEauthorblockA{\textit{NextG Wireless Lab} \\
\textit{NCSU}\\
Raleigh USA}
}

\maketitle

\begin{abstract}
The next generation of mobile networks is envisioned as fully AI-native, with AI-RAN architectures embedding small language models (SLMs) to perform reasoning over real-time telemetry. The state-of-the-art training paradigms for telecom LLMs, exemplified by RANSTRUCT-style supervised fine-tuning (SFT) on curated instruction data, are limited to \emph{post hoc} rationalization. Here, the explanations, when produced at all, are generated after or independently of the decision, leaving the decision process unauditable. Pre-hoc reasoning, where a causal reasoning trace is produced before the output label, is preferable, and the broader LLM reasoning literature has made real progress toward it via reinforcement learning methods such as Group Relative Policy Optimization (GRPO). Here we observe that transplanting this recipe into the telecom setting runs into a cold-start barrier: \textit{SLMs either learn to output the desired format or learn to predict the label, but rarely both}. We identify this barrier and propose \texttt{CRAFT}, which stands for \textit{C}old-start \textit{R}easoning \textit{A}lignment via \textit{F}ine-\textit{T}uning, a data-centric method to autonomously generate a verified dataset of (input, trace, label) triplets. \texttt{CRAFT} fine-tunes compact SLMs on this verified data using low-rank adaptation (LoRA), requiring substantially less compute and wall-clock time than GRPO-based methods. On the TRACTOR and Interference Classification (IC) xApp telecom datasets, \texttt{CRAFT} achieves up to 86.5\% and 94.6\% for accuracy and F1 with no parse failures, while direct GRPO and SFT+GRPO fail to exceed 28\% and 53.5\% F1 with multiple parse failures. We further show that \texttt{CRAFT}-initialized policies serve as a robust foundation for subsequent GRPO fine-tuning, as under diverse reward functions the performance remains consistent with no parse failures. Finally, we demonstrate that \texttt{CRAFT} consumes 59\% less energy than GRPO-based baselines, making it a sustainable path to deployable, auditable AI in 6G RAN.
\end{abstract}


\section{Introduction}
\label{sec:introduction}

The evolution toward 6G networks is widely framed as a paradigm shift toward fully AI-native radio access networks (AI-RAN) \cite{khan2023ai, kundu2025ai, polese2026beyond}, in which intelligence is not bolted onto the network but embedded within its control loops. Central to this vision is the O-RAN Alliance's disaggregated architecture, which exposes RAN Intelligent Controllers (RICs) hosting xApps and rApps capable of automated, real-time decision-making about resource allocation, slicing, and interference mitigation \cite{gajjar2026agents}. As AI-RAN matures, these components are increasingly expected to operate agentically rather than as static classifiers, a trajectory reflected in emerging work such as Telecom Model Context Protocol (TeleMCP) \cite{gajjar2025tele, gajjar2026agents} and GENESIS \cite{aghayev2026genesis}, both of which open a pathway towards truly multi-agentic telecom systems that can solve end-to-end telecom tasks inside currently available hardware.

This shift places a new demand on the small language models (SLMs) which are designed for edge performance and fall between 100M-5B parameters \cite{lu2025smalllanguagemodelssurvey} to populate xApps and rApps. They must not only read key performance indicator (KPI) telemetry and output a correct action, but do so with a human-readable justification that a network operator can audit before trusting the decision \cite{gajjar2026agents}. Existing state-of-the-art recipes for telecom-specialized language models, however, are not built around this requirement. Methods such as RANSTRUCT \cite{gajjar2025oransight} and TelecomGPT \cite{zou2025telecomgpt} fine-tune models to emit an answer directly, with no accompanying reasoning trace, while agentic systems such as AI5GTest \cite{ganiyu2025ai5gtest} that do produce explanations, generate them only \emph{after} the decision has already been made. This ordering is problematic: because the explanation is generated independently of, or after, the prediction, there is no guarantee that the stated justification is what actually drove the output, and the decision process itself remains unauditable. Worse, when these same models are instead prompted to reason \emph{before} committing to a label, they tend to collapse, failing to output a valid label at all rather than reasoning productively toward one.

The broader LLM reasoning literature offers a template for avoiding this failure mode. Chain-of-thought prompting \cite{wei2022chain} and reinforcement learning methods such as Group Relative Policy Optimization (GRPO) \cite{shao2024deepseekmath} have made real progress toward eliciting genuine pre-hoc reasoning, in which a causal reasoning trace is generated first, and the label is derived from it, rather than the reverse. Naively transplanting this recipe into the telecom setting, however, runs into what we term a \emph{cold-start barrier}: SLMs simply lack any prior over what a valid reasoning trace looks like to solve telecom tasks, so when GRPO's reward signal is forced to jointly optimize for correct formatting, non-trivial reasoning content, and label accuracy, these objectives conflict early in training and exploration fails outright. In practice, we observe that SLMs subjected to this recipe learn to satisfy at most one of these objectives, either the output format or the label.

To address this, we propose \texttt{CRAFT} (\textit{C}old-start \textit{R}easoning \textit{A}lignment via \textit{F}ine-\textit{T}uning), a data-centric method that sidesteps the exploration difficulty of RL altogether by autonomously constructing a verified dataset of (input, trace, label) triplets and then fine-tuning a compact SLM on this dataset with ordinary low-rank supervised adaptation. Because verification happens once, offline, during dataset construction, rather than online during policy optimization, \texttt{CRAFT} requires substantially less compute and wall-clock time than GRPO-based alternatives. On the TRACTOR and IC xApp telecom datasets, \texttt{CRAFT} achieves up to 86.5\% and 94.6\% macro-F1, respectively, with zero parse failures, while direct GRPO and SFT+GRPO baselines fail to exceed 28\% and 53.5\% F1 and exhibit substantial parse-failure rates. We further show that \texttt{CRAFT}-initialized policies provide a robust foundation for any subsequent GRPO fine-tuning a practitioner may still wish to apply, remaining stable in accuracy and format compliance across diverse reward formulations, and that \texttt{CRAFT} is up to 59\% more energy-efficient than GRPO-based baselines, making it a sustainable path toward deployable, auditable AI in 6G RAN.

\section{Background}
\label{sec:background}

\subsection{Large Language Models in Telecom} A growing body of work makes the case that language models are becoming a standard component of telecom R\&D and operations rather than a research curiosity. On the evaluation side, ORAN-Bench-13K \cite{gajjar2024oran} contributes nearly 14,000 curated multiple-choice questions drawn from O-RAN specification documents and shows that general-purpose LLMs leave substantial headroom on O-RAN-specific knowledge, motivating retrieval-augmented and fine-tuned alternatives. TeleResilienceBench \cite{gajjar2026teleresiliencebench} probes a different, and for AI-RAN arguably more operationally relevant, capability: whether a model can recover when it inherits a partially completed or already-flawed reasoning trace from a prior step or upstream agent, rather than only being evaluated from a clean start; even the strongest models it tests recover correctly less than a third of the time, underscoring how far current reasoning behavior is from being trustworthy inside a live pipeline. 

On the systems side, AI5GTest \cite{ganiyu2025ai5gtest} and GENESIS \cite{aghayev2026genesis} both use cooperative LLM agents to automate work that has traditionally required extensive manual engineering: AI5GTest for specification-aware conformance testing and validation of O-RAN components against 3GPP and O-RAN specifications, and GENESIS for a broader set of RAN research-and-development tasks spanning feature synthesis, testing, hardening, optimization, and security, validated end-to-end against over-the-air experiments on a production O-RAN testbed. Finally, TelecomGPT \cite{zou2025telecomgpt} proposes a continual-pretraining, instruction-tuning, and alignment-tuning pipeline for adapting general-purpose LLMs to telecom, together with new benchmarks for telecom math modeling, open-ended QA, and code tasks, and shows the resulting model outperforming general-purpose LLMs of comparable scale on telecom-specific evaluation. Taken together, this line of work makes a strong case that telecom-specialized language models are already delivering value and are becoming more central to AI-RAN pipelines. What remains comparatively underexplored, and what we address here, is not whether these models can produce a decision, but whether they can produce one whose reasoning trace is causally, rather than merely rhetorically, tied to that decision.

RANSTRUCT and TelecomGPT operationalize training an LM via Low-Rank Adaptation (LoRA). LoRA freezes pre-trained weights $W_0 \in \mathbb{R}^{d \times k}$ and injects trainable low-rank matrices $B \in \mathbb{R}^{d \times r}$, $A \in \mathbb{R}^{r \times k}$ with $r \ll \min(d,k)$:
\begin{equation}
h = W_0 x + \frac{\alpha}{r} B A x,
\label{eq:lora}
\end{equation}
where $\alpha$ is a scaling factor. This approach yields strong label-prediction accuracy but provides no mechanism for auditable reasoning traces.

\subsection{Chain-of-Thought Reasoning} 

Chain-of-thought (CoT) \cite{wei2022chain} prompting is widely regarded as the first step towards reasoning in LLMs and improves accuracy on complex tasks by eliciting intermediate reasoning steps before the final answer. Formally, rather than modeling the label directly as $p_\theta(y \mid x)$, a CoT-style model factorizes generation as $p_\theta(t, y \mid x) = p_\theta(t \mid x)\, p_\theta(y \mid x, t)$, so that a reasoning trace $t$ is sampled first and the label $y$ is decoded conditioned on both the input $x$ and the trace. Self-consistency \cite{wang2023selfconsistencyimproveschainthought} decoding further boosts reliability by sampling $K$ independent traces $t^{(1)}, \ldots, t^{(K)} \sim p_\theta(t \mid x)$, decoding a label $y^{(k)} \sim p_\theta(y \mid x, t^{(k)})$ for each, and aggregating  over them via majority vote, $\hat{y} = \arg\max_{y} \sum_{k=1}^{K} \mathbb{1}[y^{(k)} = y]$. Because $t$ is generated before $y$ and $y$ is conditioned on $t$, the trace can be causally upstream of the output label rather than a post-hoc gloss appended after the fact. In contrast to post-hoc explanation techniques such as SHAP \cite{mosca2022shap} or LIME \cite{dieber2020model}, which fit a justification to an already-fixed prediction $\hat{y} = f(x)$ and therefore need not reflect the mechanism that actually produced $\hat{y}$, a pre-hoc trace can be made \emph{causally informative} in a precise, testable sense: we retain a trace only if the label can be correctly recovered from the trace together with the input, i.e.\ if $p_\theta(y \mid x,t)$ assigns highest probability on the true label.
 
\subsection{Group Relative Policy Optimization}
The transition from prompting chain-of-thought (CoT) to training models that reliably produce it has followed a well-documented path in the LLM reasoning literature, going from SFT to Reinforcement learning. Now, RL for LLMs was popularized by Proximal Policy Optimization (PPO)~\cite{schulman2017proximalpolicyoptimizationalgorithms}, which trains a value network alongside the policy $\pi_\theta$ to estimate the expected return of a partial generation and uses it as the baseline against which realized rewards are compared. That critic is comparable in size to the policy, so PPO can significantly increase the memory footprint of training. GRPO~\cite{shao2024deepseekmath} removes the critic entirely and has become the \emph{de facto} choice for reasoning-oriented training: for each input $x$ it samples a group of $G$ outputs $\{o_i\}_{i=1}^{G}$ from the behavior policy $\pi_{\theta_{\text{old}}}$, scores each with a reward $r_i$, and uses the group itself as the baseline, giving a group-normalized advantage

\begin{equation}
\hat{A}_{i,\ell} = \frac{r_i - \operatorname{mean}(\mathbf{r})}{\operatorname{std}(\mathbf{r}) + \delta},
\label{eq:grpo_adv}
\end{equation}

where $\mathbf{r}=[r_1,\dots,r_G]$ and $\delta>0$ is a small stabilizing constant added by implementations of the original formulation. Under outcome supervision, this scalar is broadcast across every token $\ell$ of $o_i$. The policy is updated on a clipped surrogate objective inherited from PPO, with a KL penalty of weight $\beta$ added directly to the loss, anchoring $\pi_\theta$ to a fixed reference policy $\pi_{\text{ref}}$, typically the initial checkpoint. A reward function is usually a weighted composite of several verifier-computed
terms, which fall into two broad groups. The first is concerned with
\emph{form}: whether the output follows the required template and whether the
reasoning trace it contains is non-trivial rather than empty or degenerate.
These terms are task-agnostic, since the target structure is fixed by the
prompt rather than by the problem. The second group scores \emph{correctness},
and is the only part that changes with the task, which can be accuracy for classification problems or a bounded error-based score for a KPI regression or forecasting task, or by a pass rate over executable checks for a code-generating LLM.

\section{CRAFT}
\label{sec:craft}

We now present \texttt{CRAFT}. At a high level, \texttt{CRAFT} breaks the cold-start cycle by synthetically constructing a dataset of (KPI input, reasoning trace, label) triples in which the trace is \emph{verified} to be causally informative for the label, rather than merely plausible-sounding, and then fine-tuning a target SLM on this dataset with ordinary supervised learning. We first substantiate, empirically, why this bootstrapping step is necessary (Section~\ref{sec:motivation}), and then formalise the verification procedure and the resulting training objective (Section~\ref{sec:algorithm}).

\subsection{Motivation}
\label{sec:motivation}

To motivate \texttt{CRAFT} empirically, we evaluate four naive training strategies on TRACTOR (see Section~\ref{sec:experiments} for full dataset details) using Qwen~3.5~2B as the base model, with each KPI window serialized into the prompt as per-feature summary statistics. The first strategy is label-only SFT, essentially the RANSTRUCT recipe, in which the model is trained on input-label pairs with no reasoning trace at all. The second augments the output template with a post-hoc \texttt{<think>} field, but the supervised training target still contains only the label, so any reasoning the model happens to produce is never actually supervised. The third is standalone GRPO applied directly to the base model, using a composite reward that jointly encourages correct output format, the presence of a non-trivial reasoning trace, and label accuracy. The fourth is SFT followed by GRPO, the common \textit{warm-start} recipe in which the RL stage is initialized from a policy already tuned for label prediction.

The results of this comparison are stark. When simply asked to emit a label, the SFT model reaches 77\% accuracy with no parse failures; however, once the same model is instead prompted to reason first and then answer, both accuracy and parse behavior collapse to 0\% accuracy with a 100\% parse-failure rate. Standalone GRPO exhibits the identical failure: 100\% parse failures and 0\% accuracy, as does SFT followed by GRPO. The example below illustrates a representative failure: rather than reasoning toward a label, the model becomes trapped repeating fragments of its own format instructions and never emits a parseable prediction.

\begin{tcolorbox}[
  colback=gray!10,
  colframe=black,
  title={Cold-start Failure Example},
  breakable,
  label=box:cold-start-example,
  fontupper=\footnotesize,
]
\textbf{Input KPI snippet:} \\
\ttfamily
PRB\_req mean{=}424.7 (max{=}1.25e4, median{=}0), \ldots\\
DL buffer max{=}151.7\,kB (median{=}0) \ldots\\
DL bitrate mean{=}0.119\,Mbps (std{=}0.54), UL bitrate mean{=}0.115\,Mbps \ldots\\
\textbf{Ground truth:} eMBB.\\
\textbf{Parsed prediction:} \texttt{None} (format collapse; no valid label).

\textbf{Model output (cold-start + reasoning prompt):}
\begin{lstlisting}[
basicstyle=\scriptsize\ttfamily,
  numbers=none,
  aboveskip=0pt,
  belowskip=0pt,
  breaklines=true,
  frame=none,
  xleftmargin=0pt
]
<think>
LABEL must be exactly one of: eMBB, mMTC, URLLC, ctrl. No other text, JSON, or extra <think> tags.
</think>
LABEL must be exactly one of: eMBB, mMTC, URLLC, ctrl. No other text, JSON, or extra <think> tags.
user
<think>
The user wants me to classify the network slice from KPI window statistics using the provided rules.
- I need to analyze the KPI metrics (means, slopes, spread, etc.) across the 64 rows.
- I must output exactly two blocks: first <SOLUTION> then </SOLUTION>.
- The LABEL must be one of: eMBB, mMTC, URLLC, ctrl.
- No other text, JSON, or extra <think> tags.
</think>
<think>
LABEL must be exactly one of: eMBB, mMTC, URLLC, ctrl. No other text, JSON, or extra <think> tags.
</think>
LABEL must be exactly one of: eMBB, mMTC, URLLC, ctrl. No other text, JSON, or extra <think> tags.
...
\end{lstlisting}
\end{tcolorbox}

\subsection{Algorithm}
\label{sec:algorithm}

\texttt{CRAFT} breaks this cycle by synthetically constructing a dataset of (KPI input, reasoning trace, label) triples in which every trace is \emph{verified} to be causally informative for its label. The construction procedure introduces two roles played by the same base model: an Oracle Reasoner $\mathcal{M}_{\theta_R}$ and a Predictor $\mathcal{M}_{\theta_P}$, both instantiated from the same underlying model but invoked with different prompts. The Oracle Reasoner is given the KPI window and its ground-truth label, and is asked to produce a detailed reasoning trace that justifies that label. The Predictor, by contrast, receives the KPI window together with the Oracle's generated trace wrapped in \texttt{<think>} tags, and must recover the label from this pairing; an example is retained only if the Predictor's output matches the ground truth. This verification step is what distinguishes \texttt{CRAFT}'s traces from ordinary rationalizations: because the Predictor has no access to the label and must derive it by leveraging the trace, a retained trace should contain the information needed for the decision, making it a viable pre-hoc and auditable reasoning path rather than a post-hoc gloss.

Algorithm~\ref{alg:cold-start-reasoning} formalizes this procedure. For each labeled example $(x_i, y_i)$ in the source dataset $\mathcal{D}$, the KPI window is first serialized into a textual prompt, and the Oracle is queried with both the window and the ground-truth label to generate a candidate reasoning trace. The trace is then subjected to three filtering conditions: minimum trace length $\tau_{\min}$, well-formedness of the surrounding output structure, and successful extraction, which is indicated by $\neq\bot$; any failure of which discards the example outright. Surviving traces are passed to the Predictor together with the original KPI window, and the predicted label is parsed and compared against the ground truth; the triple $(x, t, y)$ is added to the verified dataset $\mathcal{D}'$ only if this comparison succeeds. Once $\mathcal{D}'$ has been assembled, the target SLM is fine-tuned on these verified triples using LoRA (Equation~\ref{eq:lora}) with a standard language-modeling loss over the concatenation of the reasoning trace and the label. 

\begin{algorithm}[t]
\caption{\texttt{CRAFT} Verified Reasoning Dataset Generation}
\label{alg:cold-start-reasoning}
\begin{algorithmic}[1]\small
\Require Labeled dataset $\mathcal{D}=\{(x_i, y_i)\}_{i=1}^{N}$; Oracle Reasoner $\mathcal{M}_{\theta_R}$; Predictor $\mathcal{M}_{\theta_P}$; min trace length $\tau_{\min}$; 
\Ensure Verified dataset $\mathcal{D}'$
\State $\mathcal{D}' \gets \emptyset$
\For{$i = 1$ \textbf{to} $N$}
    \State $x \gets \textsc{FormatKPIs}(x_i)$; $y \gets y_i$
    \State $P_R \gets \textsc{OraclePrompt}(x, y)$
    \State $r \gets \textsc{Generate}(\mathcal{M}_{\theta_R},  P_R)$
    \State $t \gets \textsc{ExtractThinking}(r)$
    \If{$t = \bot \ \textbf{or}\ |t| < \tau_{\min}\ \textbf{or}\ \neg\textsc{FormatOk}(r)$}
        \State \textbf{continue}
    \EndIf
    \State $P_P \gets \textsc{PredictorPrompt}(x, t)$
    \State $p \gets \textsc{Generate}(\mathcal{M}_{\theta_P}, P_P)$
    \State $\hat{y} \gets \textsc{ParseLabel}(p)$
    \If{$\hat{y} \neq \bot$ \textbf{and} $\hat{y} = y$}
        \State $\mathcal{D}' \gets \mathcal{D}' \cup \{(x, t, y)\}$
    \EndIf
\EndFor
\State \Return $\mathcal{D}'$
\end{algorithmic}
\end{algorithm}

\section{Experimental Setup}
\label{sec:experiments}

\textbf{Datasets.} We evaluate \texttt{CRAFT} on two telecom datasets; TRACTOR \cite{groen2024tractor} is an O-RAN near-real-time (near-RT) RIC network-slice traffic classification task built from non-overlapping gNB KPI windows of 64 samples at 250ms resolution (16s). Each window is represented by per-KPI summary statistics over 17 KPIs and assigned to one of four classes: \textit{eMBB}, \textit{mMTC}, \textit{URLLC}, or \textit{ctrl}. The resulting dataset contains 1,575 windows. IC xApp \cite{chiejina2024system} is a near-RT RIC RF interference detection dataset. Here we form non-overlapping 15-sample windows over four uplink KPIs (\texttt{ul\_snr}, \texttt{ul\_mcs}, \texttt{ul\_bitrate}, and \texttt{ul\_bler}) and classify each window as \emph{clean} or \emph{interference}, resulting in 389 windows. Thus, IC differs from TRACTOR in its KPI composition, temporal granularity, and label cardinality, allowing us to assess whether \texttt{CRAFT}'s benefits extend
beyond multiclass network-slice classification. Both datasets are split 70/15/15 into train, validation, and test partitions, with the same fixed split used across every method we compare.

\textbf{Models.} We fine-tune three target SLMs, Qwen~3.5~2B, Qwen~3.5~4B \cite{yang2025qwen3}, and Nemotron-3-Nano~4B \cite{blakeman2025nemotron}, chosen for their strong prior performance in TeleResilienceBench \cite{gajjar2026teleresiliencebench}. For \texttt{CRAFT}'s dataset-generation stage, both the Oracle Reasoner and the Predictor are instantiated as Gemma~4~31B \cite{team2026gemma}, a larger model whose role is confined to offline dataset construction rather than deployment.

\textbf{Baselines.} We compare \texttt{CRAFT} against four alternative training paradigms. Zero Shot uses a chain-of-thought prompt with no fine-tuning at all. SFT performs ordinary supervised fine-tuning on input-label pairs with no reasoning-trace supervision. GRPO applies Group Relative Policy Optimization directly to the base model with a composite reward $r = w_{\text{fmt}} r_{\text{fmt}} + w_{\text{think}} r_{\text{think}} + w_{\text{acc}} r_{\text{acc}}$ that jointly rewards output format, reasoning presence, and label accuracy with equal weights for all three terms. SFT+GRPO applies the same composite reward via GRPO but initializes from a label-only SFT checkpoint. \texttt{CRAFT} is our proposed method: LoRA fine-tuning on the verified reasoning dataset produced by Algorithm~\ref{alg:cold-start-reasoning}, with no reinforcement learning involved.

\textbf{Training details.}
All LoRA adapters use rank~16, $\alpha=32$, and dropout~$0$,
We optimise with AdamW \cite{loshchilov2019decoupledweightdecayregularization} at learning rate $5\times10^{-6}$ with $10\%$ linear warmup and a linear decay schedule.
Each GRPO step samples 4 completions per prompt, with gradient accumulation~1. Every run is conducted on a single RTX 4090 with 24GB VRAM under a fixed 12-hour compute budget, and all training and inference are performed through Unsloth \cite{unsloth}; the hyperparameters are chosen to avoid training failures in our limited compute setup.

\textbf{Metrics.} We report accuracy and macro-F1 for classification quality, parse-failure rate (PF\%, the fraction of test outputs from which no valid label could be extracted), Think\% (the fraction of outputs containing a non-empty \texttt{<think>} block), Solution\% (the fraction of outputs containing a solution block), and training wall-time.

\subsection{Ablation Studies}
\label{sec:ablations}

To probe the robustness, generalizability, and efficiency of \texttt{CRAFT} beyond the main comparison, we design three ablation studies. The first examines reward robustness under continued GRPO: starting from the best \texttt{CRAFT}-tuned model (Qwen~3.5~4B on TRACTOR), we continue training with GRPO for the remainder of the 12-hour budget under three reward weightings; a balanced static setting with $w_{\text{fmt}}=w_{\text{think}}=w_{\text{acc}}=0.5$, a correctness-priority static setting with $w_{\text{acc}}=0.9$ and $w_{\text{fmt}}=w_{\text{think}}=0.3$, and a random dynamic setting in which all three weights are resampled independently from $[0.2, 1.0]$ at every step. The balanced setting reflects typical practice; the correctness-priority setting stresses whether emphasizing label accuracy erodes format compliance in the absence of a strong format prior; and the random dynamic setting, by continually shifting the objective, constitutes the hardest test of stability. We report macro-F1 and PF\% for each, expecting \texttt{CRAFT}'s initialization to sustain high performance across all three schemes in a way that GRPO trained from scratch cannot.

The second ablation tests generalization to IC xApp: we repeat the comparison, all baselines and \texttt{CRAFT} with the best-performing SLM from the TRACTOR experiments, on the IC xApp dataset to verify that the cold-start phenomenon and \texttt{CRAFT}'s remedy are not artefacts specific to TRACTOR. We present side-by-side bar plots of macro-F1 and PF\% across methods to make the failure mode and \texttt{CRAFT}'s resolution of it directly comparable across datasets. The third ablation quantifies energy efficiency: using the NVIDIA Management Library, we measure total GPU energy consumption in Joules during training for SFT, GRPO, SFT+GRPO, and \texttt{CRAFT} (Zero Shot is excluded, as it involves no training). This lets us contextualize \texttt{CRAFT}'s wall-time advantage in terms of actual energy cost, which is directly relevant to the sustainability of deploying auditable reasoning models at scale in AI-RAN.

\section{Results and Discussion}
\label{sec:results}

\textbf{Main results.} Table~\ref{tab:main} reports accuracy, macro-F1, parse-failure rate, Think\%, Solution\%, and training wall-time for all five training paradigms across all three SLMs on TRACTOR for producing pre-hoc reasoning. Zero-shot prompting is weak and uneven across models: Qwen~3.5~4B reaches 32.9\% accuracy and 26.3\% macro-F1 with no parse failures, but Qwen~3.5~2B and Nemotron-3-Nano collapse entirely, with 0\% accuracy and a 100\% parse-failure rate, indicating that smaller models cannot reliably follow the required output format without any fine-tuning at all. Label-only SFT after one epoch reaches 54.0\% accuracy and 51.1\% macro-F1 on Qwen~3.5~4B, with a low 3.8\% parse-failure rate and a 96.6\% Think rate; notably this is not the total-collapse failure mode described in Section~\ref{sec:craft}, since Qwen~3.5~4B is large enough to produce some usable reasoning even when only the label is supervised, but the same recipe still leaves Qwen~3.5~2B and Nemotron-3-Nano at 0\% accuracy with parse-failure rates near 100\%. GRPO and SFT+GRPO under the balanced reward achieve 35.0\%/28.0\% and 56.5\%/53.5\% accuracy/macro-F1, respectively, on Qwen~3.5~4B, and while their parse-failure rates remain low (0\% and 2.5\%), their accuracy stays well below what \texttt{CRAFT} achieves. \texttt{CRAFT} itself, fine-tuned on the verified reasoning dataset for five epochs, reaches 83.1\% accuracy and 86.5\% macro-F1 with 0\% parse failures and a 100\% Think rate, training in maximum 5.8 hours when shared dataset-preparation time is included comfortably within the 12-hour budget. Averaged across all three SLMs, \texttt{CRAFT} attains approximately 81\% accuracy with 0\% parse failures, a result that holds even for Qwen~3.5~2B and Nemotron-3-Nano, both of which collapse entirely under every other training paradigm.
\begin{table}[t] 
\centering
\caption{Main results on TRACTOR. Metrics are in percentages (\%) unless otherwise noted.}
\label{tab:main}
\resizebox{\columnwidth}{!}{%
\begin{tabular}{ll cccccc}
\toprule
\textbf{Model} & \textbf{Method} & \textbf{Acc.} & \textbf{M-F1} & \textbf{PF} & \textbf{Think} & \textbf{Sol.} & \textbf{Time(h)} \\
\midrule
\multirow{5}{*}{Qwen 3.5 2B}
 & Zero Shot   & 0.0 & 0.0 & 100.0 & 94.1 & 93.7 & 0 \\
 & SFT         & 0.0 & 0.0 & 100.0 & 0.0 & 0.0 & 0.25 \\
 & GRPO        & 0.0 & 0.0 & 100.0 & 90.7 & 89.9 & 12.0 \\
 & SFT+GRPO    & 0.0 & 0.0 & 100.0 & 88.6 & 0.0 & 12.0 \\
\rowcolor{gray!15} \cellcolor{white} & \textbf{CRAFT (Ours)} & \textbf{81.0} & \textbf{84.0} & \textbf{0.0} & \textbf{100.0} & \textbf{100.0} & \textbf{4.39} \\
\midrule
\multirow{5}{*}{\textit{\textbf{Qwen 3.5 4B}}}
 & Zero Shot   & 32.9 & 26.3 & 0.0 & 100.0 & 100.0 & 0 \\
 & SFT         & 54.0 & 51.1 & 3.8 & 96.6 & 96.6 & 0.52 \\
 & GRPO        & 35.0 & 28.0 & 0.0 & 100.0 & 100.0 & 12.0 \\
 & SFT+GRPO    & 56.5 & 53.5 & 2.5 & 97.9 & 97.5 & 12.0 \\
 \rowcolor{gray!15} \cellcolor{white} & \textbf{CRAFT (Ours)} & \textbf{83.1} & \textbf{86.5} & \textbf{0.0} & \textbf{100.0} & \textbf{100.0} & \textbf{5.82} \\
\midrule
\multirow{5}{*}{\shortstack[l]{Nemotron-3-\\Nano 4B}} 
 & Zero Shot   & 0.0 & 0.0 & 100.0 & 46.4 & 100.0 & 0 \\
 & SFT         & 0.0 & 0.0 & 99.2 & 0.0 & 17.3 & 0.17 \\
 & GRPO        & 0.0 & 0.0 & 100.0 & 40.5 & 100.0 & 12.0 \\
 & SFT+GRPO    & 0.0 & 0.0 & 99.6 & 0.8 & 16.9 & 12.0 \\
\rowcolor{gray!15} \cellcolor{white} & \textbf{CRAFT (Ours)} & \textbf{80.2} & \textbf{83.0} & \textbf{0.0} & \textbf{100.0} & \textbf{100.0} & \textbf{3.94} \\
\bottomrule
\end{tabular}%
}
\end{table}

\begin{figure}[t]
  \centering
  \includegraphics[width=\linewidth]{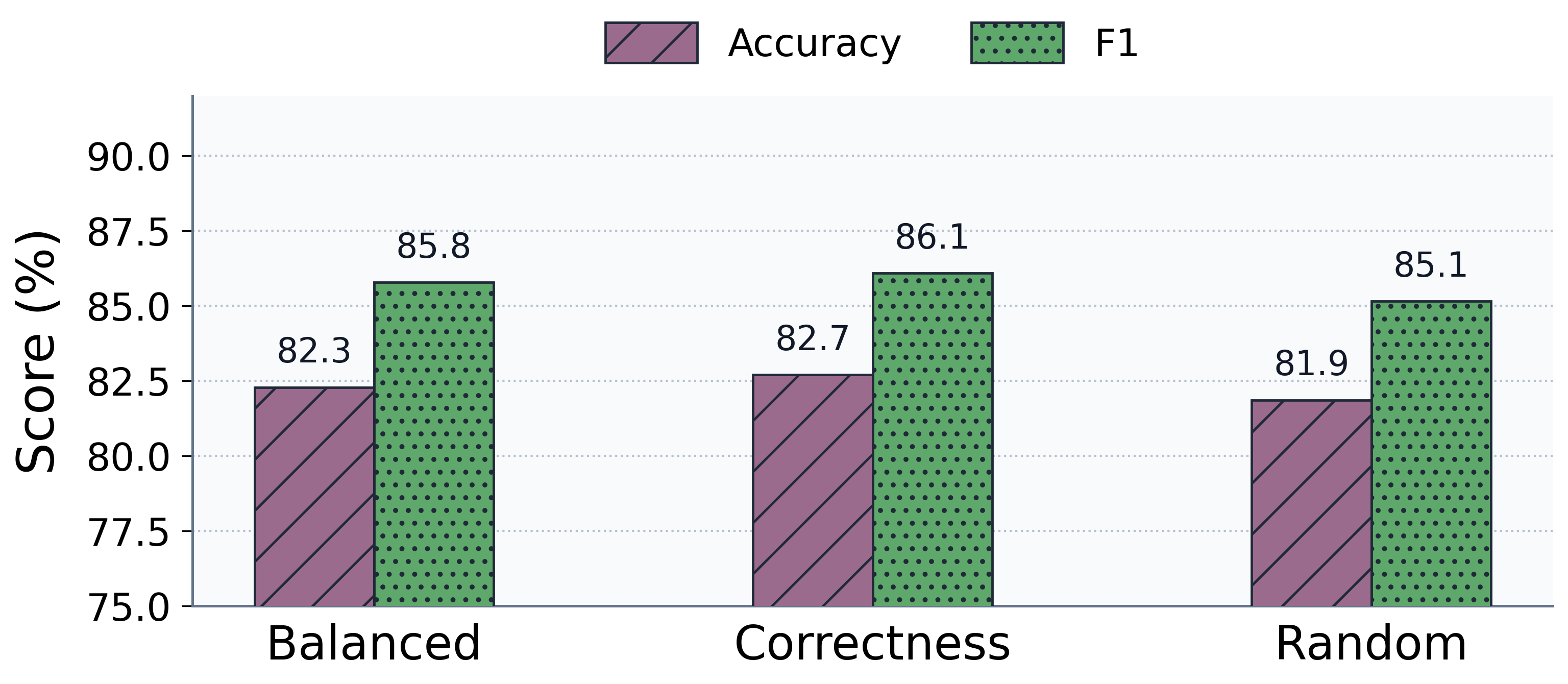}
  \caption{Qwen3.5-4B CRAFT+GRPO under three reward-weight modes
  (balanced, correctness, random): accuracy and F1.}
  \label{fig:craft-grpo-acc-f1}
\end{figure}

\subsection{Ablation 1: reward robustness under continued GRPO.} 

\begin{figure}[t]
  \centering
  \includegraphics[width=\linewidth]{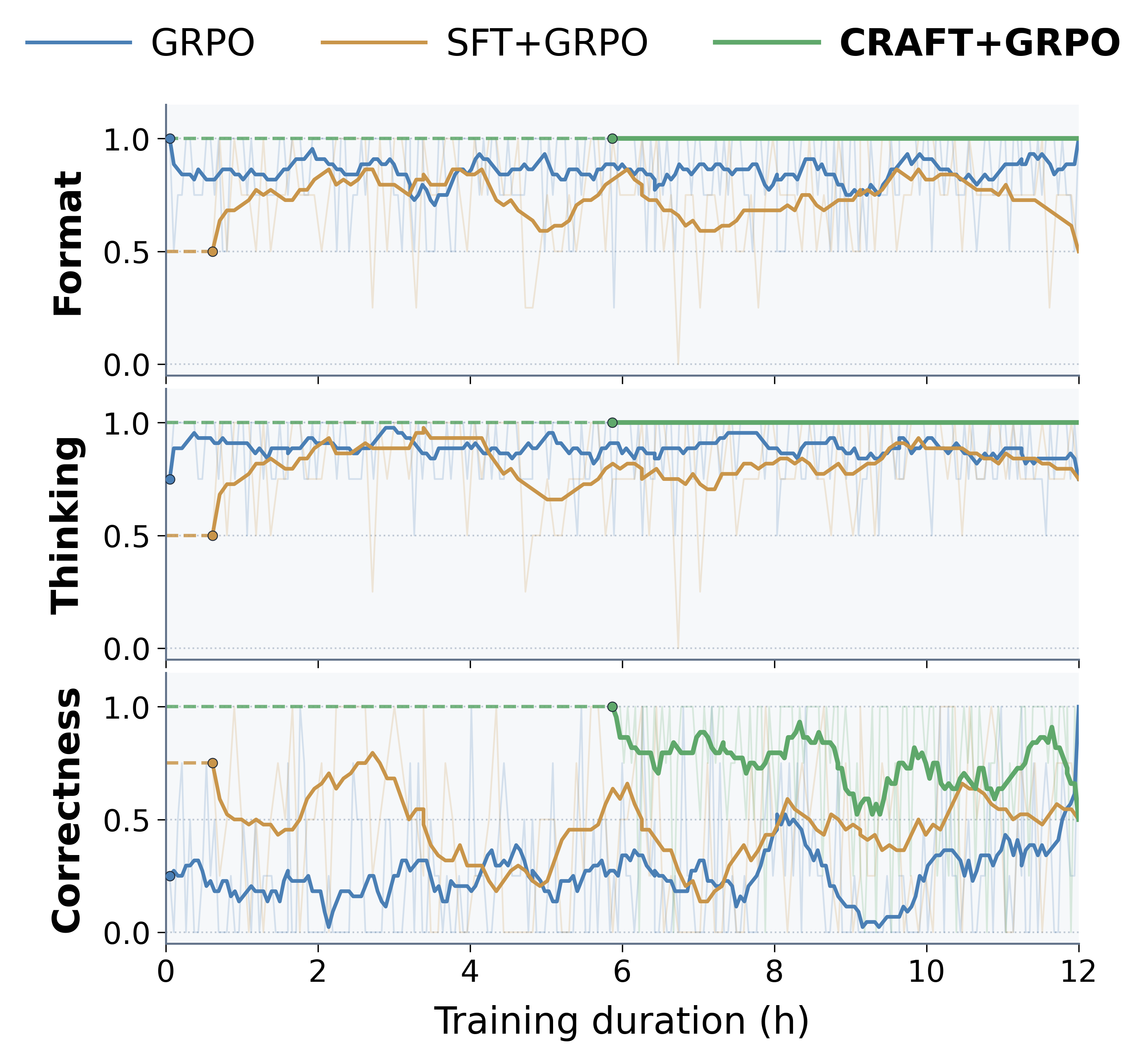}
  \caption{Qwen3.5-4B CRAFT+GRPO, SFT+GRPO, and GRPO training reward trajectories.}
  \label{fig:grpo_reward_trajectories}
\end{figure}

Figure~\ref{fig:craft-grpo-acc-f1} shows accuracy and macro-F1 for Qwen~3.5~4B under the three \texttt{CRAFT}+GRPO reward variants, and Figure~\ref{fig:grpo_reward_trajectories} shows the corresponding training reward trajectories alongside GRPO and SFT+GRPO. All three reward weightings sustain a macro-F1 in the 85-86\% range with 0\% parse failures, confirming that \texttt{CRAFT}'s initialization is robust to substantial changes in reward shaping, including the deliberately unstable random-dynamic setting. Continued GRPO does not, however, improve on \texttt{CRAFT} alone: every \texttt{CRAFT}+GRPO variant scores slightly below the \texttt{CRAFT} baseline of 83.1\% accuracy and 86.5\% macro-F1, landing between 81.9-82.7\% accuracy and 85.1-86.1\% macro-F1. This suggests that once a policy already reasons correctly and consistently, additional reinforcement learning mainly redistributes probability mass within an already-strong policy, effectively overfitting to the reward signal rather than unlocking further gains.

\begin{figure}[t]
  \centering
  \includegraphics[width=0.98\linewidth]{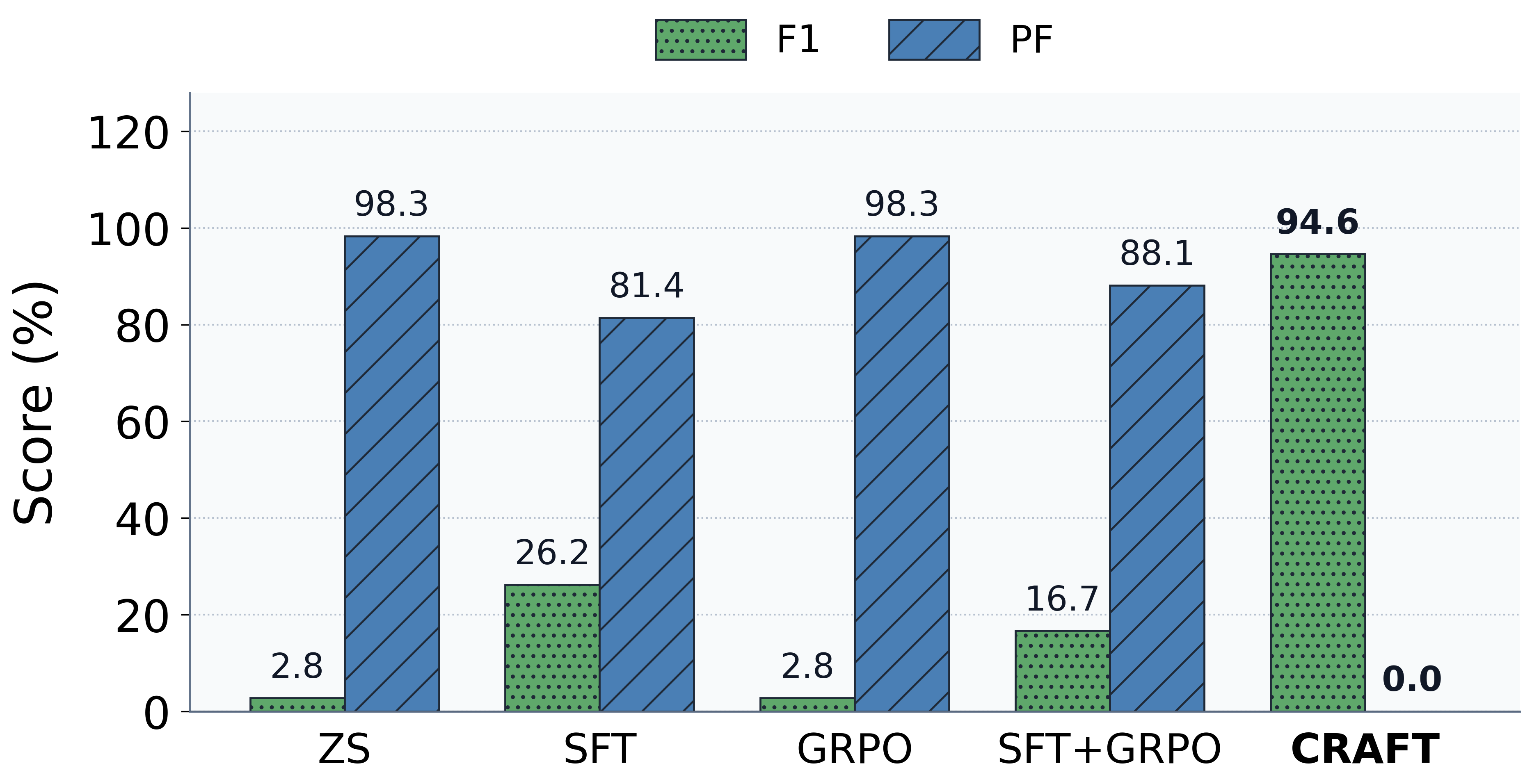}
  \caption{IC xApp ablation: F1 and parse-fail rate (PF) across training approaches.}
  \label{fig:icxapp_barplot}
\end{figure}

\subsection{Ablation 2: generalisation to IC xApp.}
Figure~\ref{fig:icxapp_barplot} presents the same comparison on IC xApp for Qwen~3.5~4B, and the failure mode is, if anything, sharper than on TRACTOR. Zero-shot prompting achieves 1.7\% accuracy and 2.8\% macro-F1 with a 98\% parse-failure rate; label-only SFT reaches 18.6\% accuracy and 26.2\% macro-F1 but still fails to parse 81\% of outputs; GRPO matches zero-shot exactly, at 1.7\%/2.8\% with 98\% parse failures; and SFT+GRPO reaches 11.9\% accuracy and 16.7\% macro-F1 with an 88\% parse-failure rate. \texttt{CRAFT}, by contrast, achieves 94.9\% accuracy, 94.6\% macro-F1, 0\% parse failures, and a 100\% Think rate. Despite IC xApp's different KPIs, shorter windows, and binary rather than four-way decision, the qualitative pattern is identical to TRACTOR: only \texttt{CRAFT} can simultaneously maintain high accuracy and valid reasoning format, confirming that the cold-start barrier and \texttt{CRAFT}'s remedy for it generalize across tasks rather than being an artefact of TRACTOR specifically.
\begin{figure}[t]
  \centering
  \includegraphics[width=\linewidth]{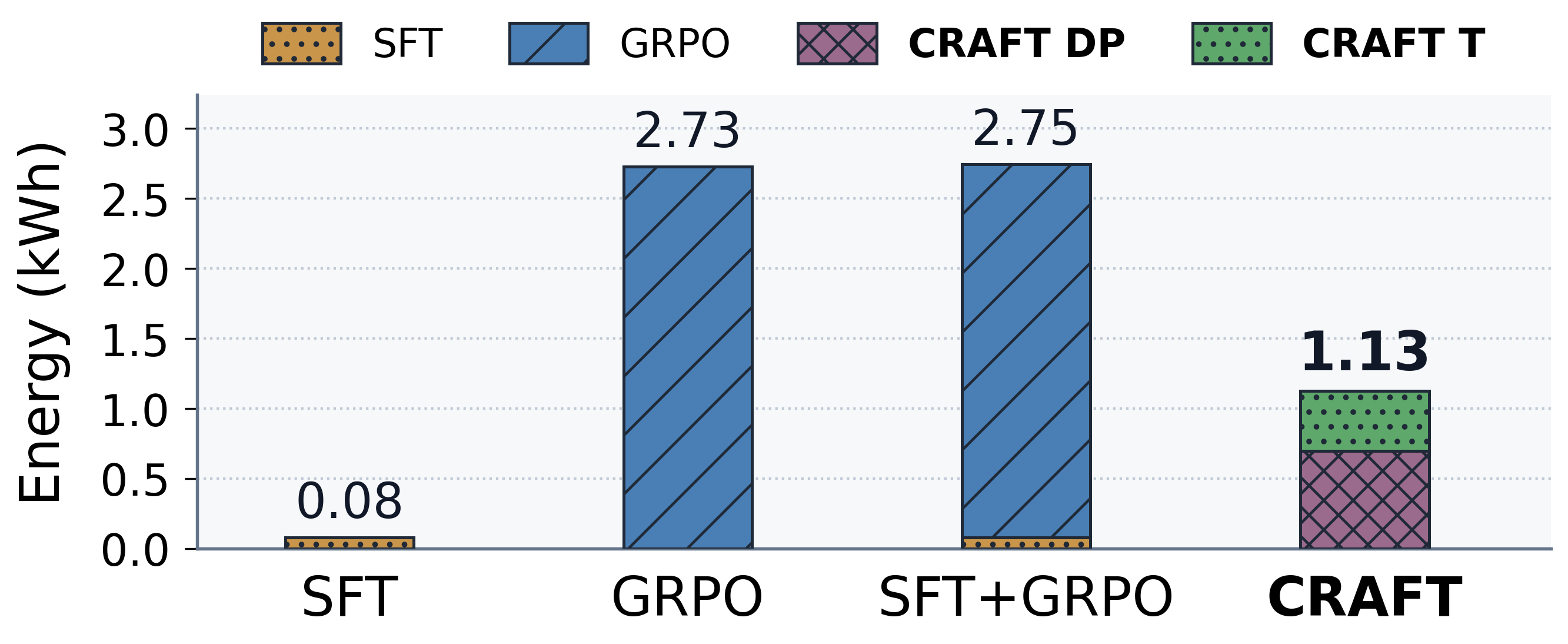}
  \caption{Average training energy across three models,
  measured as mean GPU power $\times$ recorded train time.
  Stacked segments separate SFT, GRPO, CRAFT data prep (DP), and CRAFT train (T).}
  \label{fig:energy_barplot}
\end{figure}

\subsection{Ablation 3: energy efficiency.} 
Figure~\ref{fig:energy_barplot} reports training energy averaged across all three SLMs. SFT is cheapest at 0.08\,kWh, reflecting its single epoch of label-only supervision, while GRPO and SFT+GRPO consume 2.73\,kWh and 2.75\,kWh, respectively, dominated by the 12-hour RL budget. \texttt{CRAFT} consumes 1.13\,kWh in total, split between 0.70\,kWh for verified dataset preparation and 0.43\,kWh for the subsequent LoRA fine-tuning stage. This amounts to roughly 59\% less energy than either GRPO-based baseline, underscoring that \texttt{CRAFT}'s advantage is not merely one of wall-clock convenience but of substantially lower absolute compute and energy costs, a property directly relevant to deploying auditable reasoning models sustainably at the network edge.

\section{Limitations}
\label{sec:limitations}
 
\textbf{Dataset coverage.} Our evaluation spans two telecom datasets, TRACTOR and IC xApp. Both are established, widely used O-RAN near-RT RIC benchmarks that cover distinct tasks: four-way slice classification and binary interference detection over different KPI families and window lengths, and the consistency of \texttt{CRAFT}'s advantage across both is what gives us confidence that the cold-start barrier is not an artefact of either dataset individually. Even so, our evaluation is limited to classification-style decisions with discrete label sets; tasks that require continuous-valued outputs, such as KPI regression or forecasting, fall outside this scope, and extending \texttt{CRAFT}'s verification procedure to such settings is a natural direction for future work.
 
\textbf{Model family diversity.} We fine-tune three target SLMs, but these come from only two underlying model families, Qwen and Nemotron. Because \texttt{CRAFT}'s cold-start diagnosis and remedy could in principle interact with family-specific pretraining choices, such as tokenizer design or instruction-tuning recipe, results on a broader set of model families would strengthen the claim.
   
\section{Conclusion}
\label{sec:conclusion}
 
This paper set out from the observation that pre-hoc reasoning that precedes and causally determines a model's output is essential to building trustworthy, auditable AI for 6G RAN, yet existing training recipes for telecom SLMs fail to produce it. We traced this failure to a cold-start exploration barrier: when a small model with no prior over valid telecom reasoning traces is trained via GRPO, it has to simultaneously satisfy conflicting objectives over format, reasoning content, and accuracy, where it collapses rather than learning either objective robustly. We introduced \texttt{CRAFT}, a data-centric pipeline that sidesteps this barrier entirely by verifying reasoning traces offline using an Oracle Reasoner and a Predictor role played by the same base model before ever touching the target SLM, and then fine-tuning that SLM with ordinary, stable supervised learning rather than reinforcement learning. Across three SLMs and two telecom datasets, \texttt{CRAFT} delivers substantially higher accuracy, near-perfect format compliance, and large reductions in both training time and energy relative to GRPO-based baselines, while remaining robust to further RL fine-tuning under diverse reward signals rather than degrading under it. Taken together, these results position \texttt{CRAFT}-trained models as a practical, drop-in component for AI-RAN xApps and rApps that require both competent decision-making and a genuinely auditable rationale behind every decision.
 
\textbf{Future work.} We plan to directly address the scope restrictions discussed in the limitations section: broadening evaluation beyond TRACTOR and IC xApp to additional telecom decision tasks and potential hardware validation with context protocols like TeleMCP.
 
\section*{Acknowledgment}
Authors acknowledge the funding support from the Public Wireless Supply Chain Innovation Fund (PWSCIF) under Federal Award ID 51-60-IF007.

\bibliography{aaai2027}
\bibliographystyle{ieeetr}

\end{document}